\documentclass[11pt]{article}

\usepackage[margin=1in]{geometry}
\usepackage{times}
\usepackage[T1]{fontenc}
\usepackage[utf8]{inputenc}
\usepackage{microtype}
\usepackage{amsmath,amssymb,amsthm}
\usepackage{booktabs}
\usepackage{multirow}
\usepackage{graphicx}
\usepackage{xcolor}
\usepackage{natbib}
\usepackage{hyperref}
\usepackage{cleveref}
\usepackage{algorithm}
\usepackage{algpseudocode}
\usepackage{subcaption}
\usepackage{enumitem}
\usepackage{xspace}

\hypersetup{
  colorlinks=true,
  linkcolor=blue!50!black,
  citecolor=blue!50!black,
  urlcolor=blue!50!black
}

\newcommand{\method}{PCTree\xspace}

\title{From Chains to Trees: Parent-Conditioned Drafting\\
for Semi-Autoregressive Speculative Decoding}

\author{
  Zixian Li$^{1}$ \quad
  Tong Li$^{1}$ \quad
  Chi Xie$^{1}$ \quad
  Xiaohui Song$^{1}$\thanks{Corresponding author.} \quad
  Haonan Lu$^{1}$ \\[0.4em]
  $^{1}$OPPO AI Center \\
  \texttt{lizixian@oppo.com, litong3@oppo.com, xiechi@oppo.com} \\
  \texttt{songxiaohui@oppo.com, luhaonan@oppo.com}
}

\date{}

\begin{document}
\maketitle

\begin{abstract}
Speculative decoding accelerates LLM inference only when drafted continuations survive target-model verification. Semi-autoregressive drafters such as DSpark predict an entire token block with one backbone forward and refine it with a lightweight Markov head.  However, DSpark decodes this block as a single chain, so an early mismatch invalidates the remaining suffix and limits the benefit of large draft blocks.

We show that the conditional structure already learned by DSpark can support multiple parent-consistent continuations without retraining or additional backbone passes. We introduce \textbf{P}arent-\textbf{C}onditioned Drafting \textbf{T}ree (\method), which uses the pretrained Markov head to score alternative children separately for each concrete parent and allocates a fixed verification budget to the most probable paths. This converts DSpark's linear draft into a tree while preserving its one-pass parallel backbone.

Across Qwen3-\{4B,8B,14B\} and nine benchmarks, at $B{=}7$, measured speedup gains over autoregressive (AR) decoding, relative to matched DSpark, range from $3.1\%$ to $29.5\%$.
On Qwen3-4B GSM8K at $B{=}16$, \method increases mean acceptance length from $9.41$ to $11.16$ and three-run mean AR speedup from $6.14{\times}$ to $6.60{\times}$.
These show that parent-conditioned branching can turn conditional capacity already present in a semi-autoregressive drafter into end-to-end inference gains through an inference-only change.
\end{abstract}
\section{Introduction}
\label{sec:intro}

Autoregressive decoding remains a major latency bottleneck for large language models because each step exposes little parallel work. Speculative
decoding~\citep{leviathan2023fast,chen2023accelerating,xia2023speculative} instead drafts several future tokens and verifies them together, but its benefit depends on how much of each draft survives verification. Tree proposals can
hedge against early mismatches, yet autoregressive drafters such as EAGLE~\citep{li2024eagle,li2024eagle2,li2025eagle} require multiple sequential steps to grow them~\citep{miao2024specinfer}, while parallel drafters trade stronger inter-token dependence for lower draft latency~\citep{chen2026dflash,ringel2026ddtree}.

DSpark~\citep{cheng2026dspark} combines a one-pass parallel backbone with a lightweight Markov head that conditions each draft position on its predecessor. However, its decoder uses this conditional model to construct only one chain, so a mismatch near the root invalidates the remaining suffix. This rejection cascade limits the number of tokens advanced per verification, especially for large draft blocks.

\begin{figure*}[t]
  \centering
  \IfFileExists{figures/fig1_overview.pdf}{
    \includegraphics[width=\textwidth,trim={0 8pt 0 0},clip]{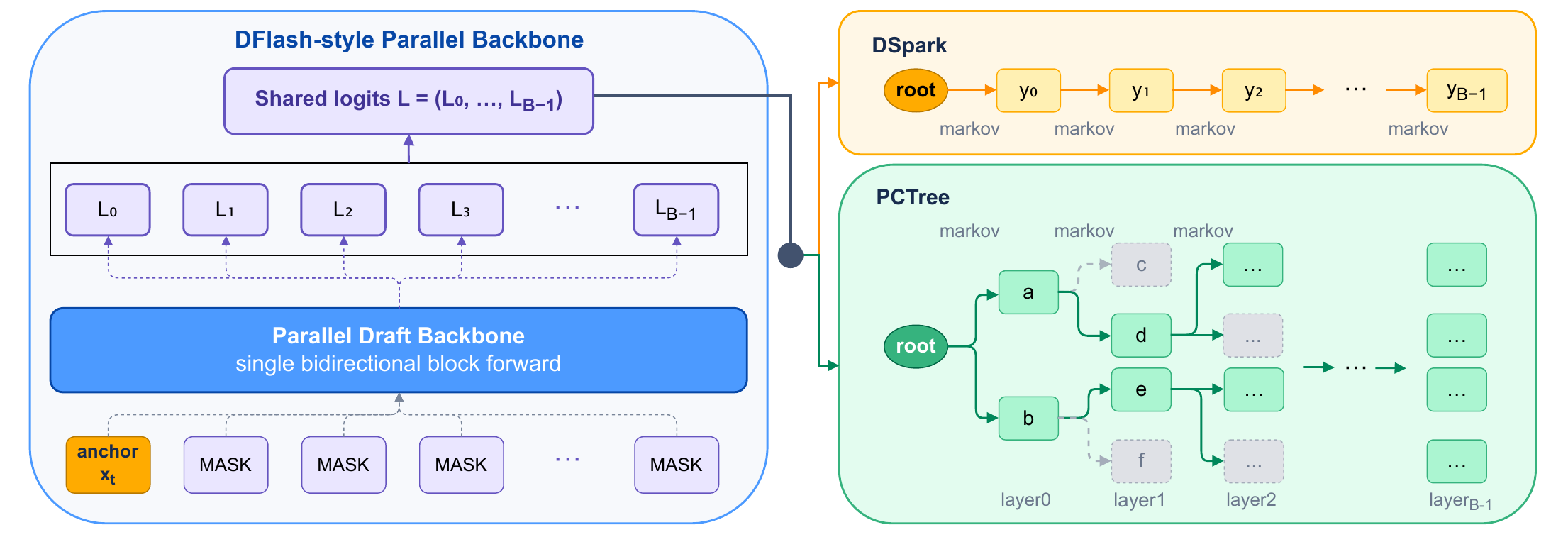}
  }{
    \fbox{\parbox[c][0.30\textheight][c]{0.96\textwidth}{
      \centering
      \Large\textbf{Figure 1 Placeholder}\\[0.8em]
      \normalsize DSpark vs.\ \method overview\\[0.5em]
      \texttt{figures/fig1\_overview.pdf}\\[0.8em]
      Export \texttt{drawio/fig1\_overview.drawio} as a cropped PDF.
    }}
  }
  \caption{\textbf{From DSpark to \method.}
  One parallel backbone forward produces shared base logits
  $L_0,\ldots,L_{B-1}$.
  DSpark forms one chain with $B$ batch-1 Markov stages, whereas
  \method batches frontier parents and retains a tree under node budget $N$.
  Both use the same pretrained weights and number of sequential stages.}
  \label{fig:overview}
\end{figure*}

Branching alone does not fully solve the problem: if all parents at one depth
share the same child distribution, individually likely tokens may form a
conditionally unlikely path.  Our key observation is that DSpark already
provides the missing signal.  We therefore introduce \method, which batches
concrete parents through the pretrained Markov head, reuses the shared backbone
logits, and retains a tree under a fixed verification budget
(\cref{fig:overview}).  This inference-only conversion requires neither
retraining nor additional backbone passes.  It does add per-round tree and
verification work, which must be offset by accepting longer paths in fewer
rounds.

Experiments show that this trade-off is favorable in the evaluated settings.
A matched Shared-Markov control attributes longer acceptance to parent-specific
conditioning rather than branching alone.  Across Qwen3-\{4B,8B,14B\} and nine
benchmarks at $B{=}7$, \method improves acceptance in every model--task pair
and raises measured end-to-end AR speedup by $3.1\%$--$29.5\%$ over DSpark.
On Qwen3-4B GSM8K at $B{=}16$, it increases mean acceptance length from $9.41$
to $11.16$ and three-run mean AR speedup from $6.14{\times}$ to $6.60{\times}$.

The paper makes three contributions:
\begin{itemize}[leftmargin=1.2em]
  \item \textbf{A decoding perspective on semi-autoregressive drafting.}
  We show that DSpark's parent-dependent factorization can support tree-shaped
  proposals despite its original single-chain decoder.
  \item \textbf{Training-free parent-conditioned tree drafting.}
  We propose \method, which constructs a fixed-budget tree to improve acceptance length without changing the pretrained
  drafter.
  \item \textbf{Controlled mechanism and end-to-end evidence.}
  We show the importance of parent conditioning and the gain in acceptance and wall-clock through experimental evidence.
\end{itemize}

\section{Related Work}
\label{sec:related}

\paragraph{Speculative decoding and tree verification.}
Early speculative decoding systems draft a single continuation and ask the target model to accept a prefix of that continuation~\citep{leviathan2023fast,chen2023accelerating,xia2023speculative}.
Because later tokens are discarded whenever an earlier one fails, chain drafts are brittle.
Tree-based verification~\citep{miao2024specinfer} addresses this fragility by packing many alternative continuations into one attention pattern, so that a rejected sibling does not invalidate an entire parallel branch.
Follow-up work improved the \emph{drafter} that feeds such trees: multi-head proposals~\citep{cai2024medusa}, feature-level drafting with target hidden states~\citep{li2024eagle}, dynamic tree shapes guided by draft confidence~\citep{li2024eagle2,wang2025opt}, and multi-layer feature fusion~\citep{li2025eagle}.
These lines largely assume that drafting itself is autoregressive, so growing a tree costs additional draft forwards.

\paragraph{Parallel drafting.}
Parallel generation reduces draft latency by predicting several positions together. Non-autoregressive models remove within-sequence dependencies to generate all positions in parallel~\citep{gu2018nonautoregressive}, whereas semi-autoregressive models generate groups in parallel while retaining dependencies between successive groups~\citep{wang2018semi}.
Blockwise parallel decoding similarly predicts several future positions and validates the longest matching prefix~\citep{stern2018blockwise}.
Other related approaches include shared-trunk multi-token prediction~\citep{gloeckle2024multitoken} and multi-head speculative drafting~\citep{cai2024medusa}.
\citet{chen2026dflash} introduced DFlash, a lightweight block diffusion drafter that emits an entire draft block in one forward while conditioning on target-side context features.
Vanilla DFlash still verifies essentially one drafted trajectory per round.

\paragraph{Trees from parallel drafts.}
\citet{ringel2026ddtree} extended this approach with DDTree, which constructs a draft tree from the per-position distributions of a block diffusion
drafter and selects nodes under a fixed budget with a best-first procedure.
\citet{wang2026taps} propose TAPS, which estimates path-level reach probabilities with a lightweight target-aware scorer and selects a prefix-closed subtree under a verification budget.

\paragraph{Semi-autoregressive speculative decoding.}
In DSpark, \citet{cheng2026dspark} argue that pure parallel blocks suffer from rapid acceptance decay because they under-model intra-block dependencies, while fully autoregressive drafters remain sequential.
It therefore uses a parallel backbone plus a cheap sequential module (the Markov head) and further introduces confidence-scheduled verification for serving.
The original DSpark decoding path, however, still materializes a linear draft for verification.
\method instead uses DSpark's pretrained Markov head to condition each expansion on its parent, without additional training.
\Cref{tab:vs_ddtree} contrasts this factorization with DSpark and DDTree.

\begin{table}[t]
\centering
\caption{Conceptual comparison of parallel-block drafting strategies.
DSpark and \method share the same parallel--Markov drafter but construct a
single chain and a budgeted tree, respectively.}
\label{tab:vs_ddtree}
\small
\resizebox{\linewidth}{!}{%
\begin{tabular}{lcccc}
\toprule
 & DFlash & DDTree & DSpark & \method \\
\midrule
Draft model & block diffusion & block diffusion & parallel+Markov & parallel+Markov \\
Candidate structure & chain & budgeted tree & chain & budgeted tree \\
Score source & per-position $p_d$ & per-position $p_d$ & Markov-refined logits & Markov-refined logits \\
Parent-specific scoring & no & no & one active parent & \textbf{batched parents} \\
Extra training for tree & -- & no & -- & \textbf{no} \\
Seq.\ draft stages / round & $1$ backbone & $1$ backbone & $1$ backbone + $B$ Markov & $1$ backbone + $B$ Markov \\
\bottomrule
\end{tabular}
}
\end{table}

\section{Preliminaries}
\label{sec:background}

\subsection{Speculative Decoding}
\label{sec:prelim_sd}

Speculative decoding uses a lightweight draft model $\mathcal{M}_D$ to propose a block of future tokens and a target model $\mathcal{M}_T$ to verify the entire block in one forward pass~\citep{leviathan2023fast,chen2023accelerating}.
Verification proceeds from left to right and commits the longest prefix consistent with the target distribution; after the first rejection, the remaining draft suffix is discarded and a target-generated bonus token completes the round.
Its efficiency is governed by the trade-off among draft latency, target-verification latency, and the number of tokens advanced per round.
Draft architectures therefore seek to raise acceptance without making candidate generation or verification disproportionately expensive.

\subsection{DSpark: Semi-Autoregressive Drafting}
\label{sec:prelim_dspark}

We next recall the chain-style DSpark decoder~\citep{cheng2026dspark} that our method generalizes.
Let $\mathcal{M}_D$ be a DSpark draft model with block size $B$ and let
$x_t$ be the last verified token. Drafting has a parallel backbone stage
followed by a lightweight sequential stage.

\paragraph{Parallel backbone.}
The DFlash-style backbone~\citep{chen2026dflash} processes the anchor token
$x_t$ and $B{-}1$ mask placeholders in one forward pass, producing base
logits $L=(L_0,\ldots,L_{B-1})$ for $B$ draft positions. Because the
positions are computed in parallel, $L_d$ does not depend on the concrete
tokens selected earlier in the block.

\paragraph{Sequential Markov head.}
DSpark restores within-block dependence with a low-rank Markov head that adds
a bias determined by the preceding token. Starting from $y_{-1}:=x_t$, it
applies the head sequentially for $d=0,\ldots,B-1$:
\begin{equation}
  \label{eq:markov_step}
  z_d = L_d + \mathrm{Markov}(y_{d-1}),\qquad
  y_d = \arg\max_{v\in\mathcal{V}}\; \mathrm{softmax}(z_d)_v.
\end{equation}
This produces one chain $(y_0,\ldots,y_{B-1})$: one backbone forward followed
by $B$ sequential batch-1 Markov calls (\cref{fig:overview}).

\paragraph{Scope.}
DSpark also considers an RNN head that maintains the full within-block prefix
state, but uses the Markov head by default because the RNN head provides only
marginal gains at higher implementation complexity~\citep{cheng2026dspark}.
We follow this default; the Markov head's first-order state also permits
batched scoring over multiple parents.

\section{\method}
\label{sec:method}

The top-$1$ choice in \cref{eq:markov_step} produces a single chain.
\method replaces this choice with budgeted branching while reusing the same
backbone logits, Markov operator, and pretrained parameters.

\begin{figure*}[t]
  \centering
  \IfFileExists{figures/fig3_parent_conditioning.pdf}{
    \includegraphics[width=\textwidth,trim={18pt 5pt 18pt 0},clip]{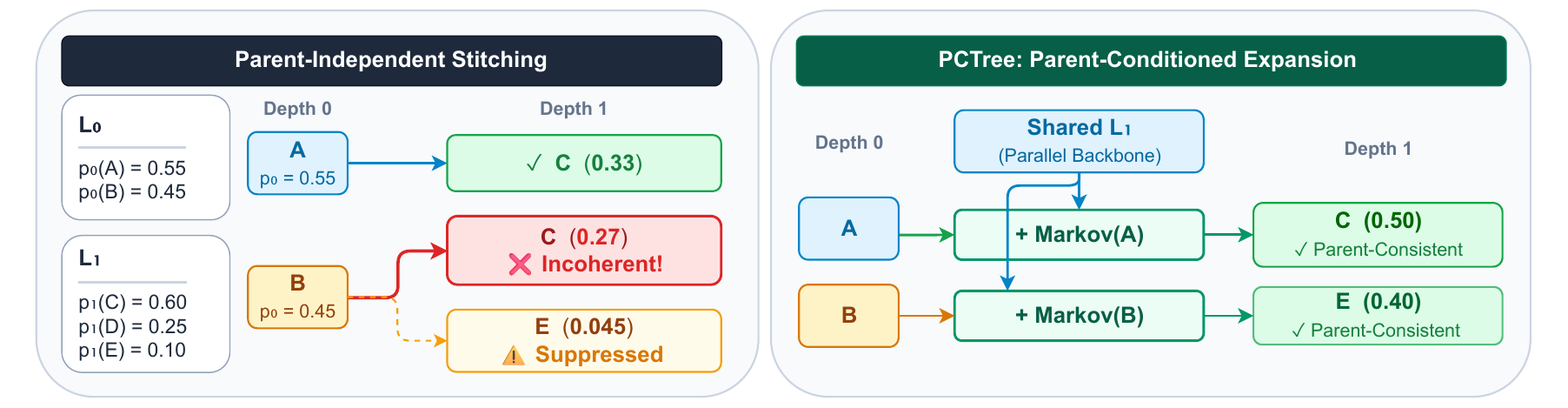}
  }{
    \fbox{\parbox[c][0.28\textheight][c]{0.96\textwidth}{
      \centering
      \Large\textbf{Figure 3 Placeholder}\\[0.8em]
      \normalsize Marginal stitching vs.\ parent-conditioned expansion\\[0.5em]
      \texttt{figures/fig3\_parent\_conditioning.pdf}\\[0.8em]
      Export \texttt{drawio/fig3\_parent\_conditioning.drawio} as a cropped PDF.
    }}
  }
  \caption{\textbf{Parent-independent stitching versus parent-conditioned
expansion.}
Reusing one depth-wise distribution can favor the incoherent path
$B{\rightarrow}C$ and suppress $B{\rightarrow}E$ (left).
\method reconditions the shared logits on each parent, recovering
parent-consistent branches (right).
Values are illustrative joint path probabilities.}
  \label{fig:parent-conditioning}
\end{figure*}

\paragraph{Failure mode of parent-independent trees.}
Parallel block drafters produce one distribution for each future position in
a single forward pass~\citep{chen2026dflash}.  A tree can be constructed
directly from these depth-wise distributions, as in DDTree
~\citep{ringel2026ddtree}.  Without an additional parent-conditioned
refinement, every parent at depth $d{-}1$ is expanded with the same
distribution $q_d$; we refer to this as \emph{parent-independent
per-position scoring}.
\Cref{fig:parent-conditioning} illustrates its possible failure mode:
individually likely tokens can be combined into a locally incoherent path.
In the example, the high position-wise score of $C$ causes
$B{\rightarrow}C$ to outrank the parent-consistent continuation
$B{\rightarrow}E$, even though $C$ is implausible after $B$.
The probabilities are illustrative.

DSpark already provides the missing conditional signal: its Markov head can
re-score the shared logits for each concrete parent.  We use this signal to
construct the tree described next.

\subsection{Markov Tree Expansion}
\label{sec:expansion}

Let $k$ be the local branching factor and $N$ the verification node budget (including the root).
We index Markov stages by $d=0,\ldots,B-1$; stage $d$ constructs non-root
tree level $d+1$. Thus, block size $B$ is also the maximum number of drafted
tree levels.
Let $\mathcal{F}_{d-1}$ be the parent frontier used at stage $d$, with
$\mathcal{F}_{-1}=\{x_t\}$ at the root.
Its batch size is $b_d=|\mathcal{F}_{d-1}|$, with $b_0=1$ and
$1\le b_d\le k$.
Every parent $p\in\mathcal{F}_{d-1}$ shares the same base logits $L_d$ but receives its own Markov bias:
\begin{equation}
  \label{eq:tree_step}
  z_d(p) = L_d + \mathrm{Markov}(p),\qquad
  \log \pi_d(\cdot\mid p) = \log\mathrm{softmax}\bigl(z_d(p)\bigr).
\end{equation}
Equivalently, stack the frontier tokens into
$P_{d-1}=(p_1,\ldots,p_{b_d})\in\mathcal{V}^{b_d}$.
At sequential Markov stage $d$, one call batches the current $b_d$ parents and computes
\begin{equation}
  \label{eq:tree_step_batch}
  Z_d
  =
  \mathbf{1}_{b_d}L_d^\top
  + \mathrm{Markov}(P_{d-1})
  \in\mathbb{R}^{b_d\times|\mathcal{V}|},
\end{equation}
where the shared $L_d$ is broadcast across the $b_d$ frontier parents.

\begin{figure*}[t]
  \centering
  \IfFileExists{figures/fig2_tree_construction.pdf}{
    \includegraphics[width=\textwidth,trim={15pt 6pt 15pt 0},clip]{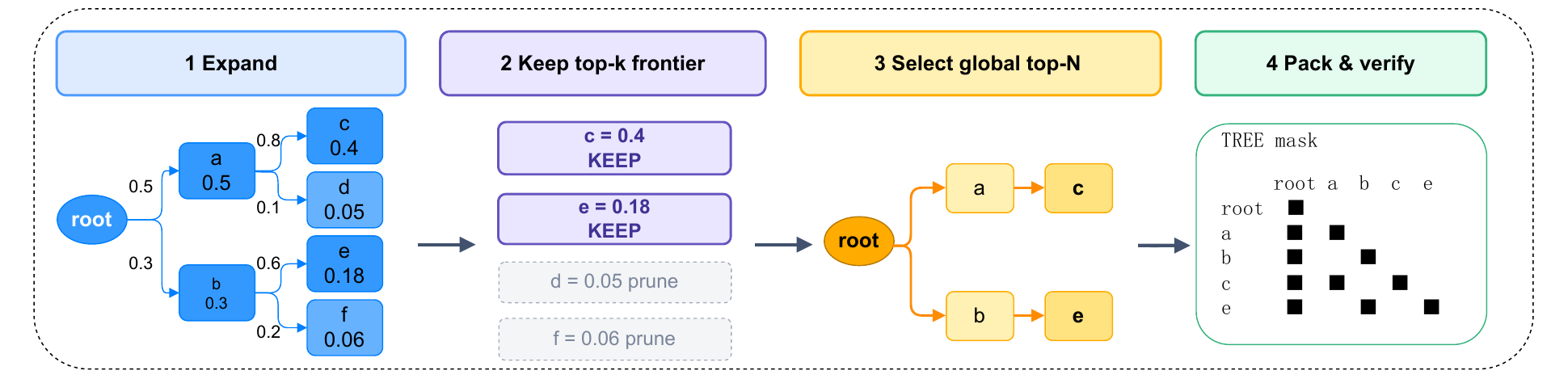}
  }{
    \fbox{\parbox[c][0.28\textheight][c]{0.96\textwidth}{
      \centering
      \Large\textbf{Figure 2 Placeholder}\\[0.8em]
      \normalsize Tree expansion, pruning, global selection, and verification\\[0.5em]
      \texttt{figures/fig2\_tree\_construction.pdf}\\[0.8em]
      Export \texttt{drawio/fig2\_tree\_construction.drawio} as a cropped PDF.
    }}
  }
  \caption{\textbf{Budgeted tree construction and verification.}
  Each frontier parent proposes top-$k$ Markov-conditioned children;
  layer-wise pruning controls expansion, and global top-$N$ selection yields
  an ancestor-closed tree.
  The selected nodes are packed with an ancestor-only mask for one target
  verification.}
  \label{fig:tree-construction}
\end{figure*}

Each candidate node $c$ is ranked by its \emph{joint path score}
\begin{equation}
  \label{eq:joint}
  s(c)=\sum_{(u\rightarrow v)\in\mathrm{path}(c)}
  \log \pi\bigl(v\mid u\bigr).
\end{equation}
We take the local top-$k$ children of each parent under
$\log\pi_d(\cdot\mid p)$ and append them to a candidate pool with updated scores
$s(\mathrm{child})=s(p)+\log\pi_d(\mathrm{child}\mid p)$.
Without pruning, the frontier entering stage $d$ would contain up to $k^d$
parents, and expanding it would produce up to $k^{d+1}$ nodes. This
exponential growth is prohibitive for large block sizes.
Following the score-guided dynamic-tree principle of EAGLE-2~\citep{li2024eagle2}, we allocate expansion capacity to the most promising current nodes: after adding all children to the pool, only the $k$ candidates with the highest joint scores become the next frontier
(\cref{fig:tree-construction}).
The frontier is therefore bounded by $k$, and the number of candidate nodes
added to the pool is at most
$k+(B-1)k^2$ rather than growing exponentially with depth.
This bound matters especially for parallel drafters with large $B$.
Without layer-wise pruning, the number of active parents would grow rapidly,
and all of their previous-token IDs would enter the next batched Markov call.
An overly wide frontier would therefore increase both the candidate count and
the latency of every subsequent Markov stage.
Thus the Markov head is still called once per depth---$B$ calls in total---but its batch size is $b_d\in\{1,\ldots,k\}$ (typically $k$ after the first step), rather than always~$1$.

\paragraph{Relation to dynamic trees and beam search.}
Frontier pruning is beam-like, but it controls only which nodes are expanded.
Unlike beam search, which returns terminal sequences, we retain scored
internal nodes because they are themselves verification candidates and
ancestors of deeper nodes. This follows EAGLE-2's score-guided dynamic-tree
principle~\citep{li2024eagle2}, but uses parent-conditioned Markov scores over
shared block logits.

\subsection{Budgeted Selection and Tree Verification}
\label{sec:selection}
\label{sec:verify}

After $B$ stages, we rank all candidate-tree nodes, including the root, by
decreasing joint score, breaking ties by shallower depth and then a stable
node identifier, and retain the global top $N$. The root has score
$s(x_t)=0$ and is therefore always selected. Because
$s(c)=s(p)+\log\pi(c\mid p)\le s(p)$, every ancestor precedes its descendants,
so the selected nodes form a prefix-closed tree~\citep{li2024eagle2}.

The resulting set induces: (1) a flat draft-token tensor for the target; (2) retrieve indices enumerating root-to-leaf paths; (3) an ancestor-only tree attention mask for a single target forward.
The packed tree is then verified in one target-model forward using the
standard greedy tree speculative-decoding rule~\citep{miao2024specinfer,li2024eagle2}.

\Cref{alg:dspark_tree} summarizes the procedure.
When $k{=}1$, expansion collapses to the unique greedy child at every depth and recovers DSpark (up to the global budget $N$).

\begin{algorithm}[t]
\caption{\method draft construction (training-free)}
\label{alg:dspark_tree}
\begin{algorithmic}[1]
\Require Draft model $\mathcal{M}_D$, root token $x_t$, block size $B$, top-$k$, budget $N$
\State $L \gets$ one parallel-block forward of $\mathcal{M}_D$ on $(x_t,\mathrm{MASK},\ldots)$
\State $\mathrm{frontier}\gets\{x_t\}$ with joint score $0$; $\mathrm{pool}\gets\{x_t\}$
\For{$d=0$ to $B-1$}
  \State Sequential stage $d$: batch all $p\in\mathrm{frontier}$ in one Markov call and compute $z_d(p)\gets L_d+\mathrm{Markov}(p)$
  \State Expand local top-$k$ children; push into $\mathrm{pool}$ with updated joint scores~\eqref{eq:joint}
  \State $\mathrm{frontier}\gets$ top-$k$ nodes of the current layer by joint score
\EndFor
\State $S\gets$ global top-$N$ pool nodes by (score desc., depth asc., stable ID)
\State \Return draft tokens, retrieve indices, and tree attention mask induced by $S$
\end{algorithmic}
\end{algorithm}

\subsection{Why Trees Help under Shared Logits}
\label{sec:why}

A useful property of DSpark is that siblings at the same depth share $L_d$ and differ only through the Markov bias of their parents.
Alternative branches are cheaper than repeating the parallel backbone:
once $L_d$ is available, expansion needs a batched Markov evaluation and
top-$k$ selection rather than another backbone forward.
This additional draft cost remains small relative to target verification and
can be amortized by longer accepted paths and fewer proposal rounds.
An autoregressive drafter, by comparison, typically pays a full draft
forward per depth for every active parent, or an equivalent batched tree
forward with growing masks.
DSpark's existing factorization therefore supports tree expansion without a
new drafter architecture.

\subsection{Complexity}
\label{sec:complexity}

Let $C_{\mathrm{bb}}$ be the cost of one backbone forward on $B$ positions and $C_{\mathrm{mk}}(b)$ the cost of a Markov call with batch $b$.
DSpark costs roughly $C_{\mathrm{bb}}+\sum_{d=0}^{B-1} C_{\mathrm{mk}}(1)$.
\method costs $C_{\mathrm{bb}}+\sum_{d=0}^{B-1} C_{\mathrm{mk}}(b_d)$ with $b_0{=}1$ and $b_d\le k$ thereafter, plus negligible selection overhead.
Thus tree construction preserves the $B$ sequential stages and increases
only their Markov batch sizes, bounded by $k$.
\section{Experiments}
\label{sec:exp}

\method changes only the inference policy. Within each matched comparison,
DSpark and \method use the same draft checkpoint, with no additional training for tree expansion.

\subsection{Setup}
\label{sec:setup}

\paragraph{Models.}
We evaluate Qwen3-\{4B,8B,14B\}~\citep{yang2025qwen3}.
The $B{=}7$ experiments use the official DSpark checkpoints released by
DeepSeek. For $B{=}16$, we train target-matched drafts with the official
DSpark recipe for each target size. Checkpoint and training details are given
in \cref{app:setup}.

\paragraph{Baselines.}
DSpark~\citep{cheng2026dspark} is the matched control and differs from
\method only in chain versus tree expansion.
EAGLE-3~\citep{li2025eagle} and DFlash~\citep{chen2026dflash} are published
references reproduced from \citet{cheng2026dspark}, not strict
same-environment speed comparisons.

\paragraph{Decoding hyperparameters.}
All matched main comparisons use a fixed reference configuration with local branching factor $k{=}4$ and verification budget $N{=}32$ (including the root).
This reference budget is held constant across models and benchmarks and is not claimed to maximize throughput on every workload.
The chain baseline is \textbf{DSpark} only; $k{=}1$ is mathematically the same greedy Markov chain and is not evaluated as a second baseline.
We report greedy target verification.
Confidence scheduling is disabled, so every comparison uses the same fixed
verification policy.

\paragraph{Metrics.}
Following \citet{cheng2026dspark}, mean acceptance length $\tau$ includes the
target bonus token.  We also report end-to-end speedup over standard
autoregressive target-only decoding and phase-wise latency.
The AR baseline uses the target model alone with HuggingFace generation, producing one token per decoding step without speculative drafting.

\paragraph{Hardware.}
All latency and AR-speedup measurements use a single NVIDIA H20 with bfloat16 and SDPA attention.
Additional checkpoint identifiers are listed in \cref{app:setup}.

\paragraph{Measurement protocol and uncertainty.}
\label{sec:measurement}
The Qwen3-4B $B{=}16$ main configurations are averaged over three full runs.
For the smallest original margin, Qwen3-4B on GSM8K at $B{=}7$, we likewise
ran DSpark and \method three times with the same prompts, prompt order,
generated-token cap, checkpoint, and decoding settings.
Other main-table configurations currently have one full run, so their decimal
precision should not be interpreted as statistical confidence.
The aggregate results appear in \cref{tab:gain}; the individual repeated
runs are reported in \cref{tab:repeatability} in the appendix.
Each point in the verification-budget sweep is likewise repeated three times.
Timing, warmup, and execution-order details are provided in
\cref{app:protocol}.

\paragraph{Benchmarks.}
Following the DSpark evaluation protocol~\citep{cheng2026dspark}, we cover
\textbf{Math}: GSM8K~\citep{cobbe2021gsm8k}, MATH-500~\citep{lightman2024verify}, and AIME25 as distributed by MathArena~\citep{balunovic2025matharena};
\textbf{Code}: MBPP~\citep{austin2021program}, HumanEval~\citep{humaneval}, and LiveCodeBench (LCB)~\citep{jain2025livecodebench};
\textbf{Chat}: MT-Bench~\citep{mtbench}, Alpaca~\citep{taori2023alpaca}, and Arena-Hard~\citep{li2025arenahard}.

\subsection{Main Results: Acceptance Length}
\label{sec:main}

\Cref{tab:main} reports mean acceptance length $\tau$ for all drafters.
\method improves $\tau$ across all nine benchmarks for each evaluated target/block-size setting.
\Cref{tab:gain} reports the corresponding Qwen3-4B acceptance and AR-speedup
gains on three representative tasks, including the three-run $B{=}16$
measurements and the repeated GSM8K measurement at $B{=}7$.
AR-speedup results for all models and benchmarks are deferred to
\cref{tab:all_throughput} in the appendix.

\begin{table*}[t]
\centering
\caption{Mean acceptance length $\tau$.
$^\dagger$: published values reproduced from \citet{cheng2026dspark},
not strict same-environment comparisons.
Non-dagger rows are our matched runs; bold marks \method.}
\label{tab:main}
\setlength{\tabcolsep}{2.8pt}
\small
\resizebox{\textwidth}{!}{%
\begin{tabular}{llccccccccc}
\toprule
 & & \multicolumn{3}{c}{Math} & \multicolumn{3}{c}{Code} & \multicolumn{3}{c}{Chat} \\
\cmidrule(lr){3-5}\cmidrule(lr){6-8}\cmidrule(lr){9-11}
Target & Drafter & GSM8K & MATH-500 & AIME25 & MBPP & HumanEval & LCB & MT-Bench & Alpaca & Arena-Hard \\
\midrule
\multirow{6}{*}{Qwen3-4B}
 & EAGLE-3$^\dagger$ & 5.14 & 4.62 & 3.92 & 3.69 & 4.16 & 3.77 & 2.39 & 2.26 & 2.55 \\
 & DFlash$^\dagger$ & 5.40 & 4.85 & 4.15 & 4.40 & 4.74 & 4.18 & 3.07 & 2.96 & 2.83 \\
 & DSpark ($B{=}7$) & 6.31 & 6.24 & 5.47 & 5.33 & 5.60 & 5.30 & 3.82 & 3.67 & 3.81 \\
 & \textbf{\method} \textbf{($B{=}7$)} & \textbf{7.24} & \textbf{7.22} & \textbf{6.77} & \textbf{6.53} & \textbf{6.78} & \textbf{6.30} & \textbf{5.07} & \textbf{4.83} & \textbf{4.83} \\
 & DSpark ($B{=}16$) & 9.41 & 9.03 & 6.87 & 6.90 & 7.42 & 6.71 & 4.28 & 3.99 & 4.30 \\
 & \textbf{\method} ($B{=}16$) & \textbf{11.16} & \textbf{10.85} & \textbf{8.80} & \textbf{8.43} & \textbf{9.21} & \textbf{8.05} & \textbf{5.50} & \textbf{5.21} & \textbf{5.44} \\
\midrule
\multirow{6}{*}{Qwen3-8B}
 & EAGLE-3$^\dagger$ & 5.30 & 4.77 & 3.91 & 3.96 & 4.33 & 4.17 & 2.66 & 2.54 & 2.54 \\
 & DFlash$^\dagger$ & 5.33 & 4.91 & 4.07 & 4.36 & 4.64 & 4.39 & 3.11 & 2.98 & 2.81 \\
 & DSpark ($B{=}7$) & 6.46 & 6.35 & 5.57 & 5.46 & 5.85 & 5.48 & 3.87 & 3.77 & 3.88 \\
 & \textbf{\method} \textbf{($B{=}7$)} & \textbf{7.36} & \textbf{7.28} & \textbf{6.77} & \textbf{6.68} & \textbf{6.96} & \textbf{6.51} & \textbf{4.98} & \textbf{4.93} & \textbf{4.91} \\
 & DSpark ($B{=}16$) & 9.75 & 9.13 & 6.93 & 6.97 & 7.88 & 6.80 & 4.35 & 4.07 & 4.37 \\
 & \textbf{\method} ($B{=}16$) & \textbf{11.50} & \textbf{10.93} & \textbf{8.74} & \textbf{8.76} & \textbf{9.65} & \textbf{8.28} & \textbf{5.63} & \textbf{5.40} & \textbf{5.52} \\
\midrule
\multirow{6}{*}{Qwen3-14B}
 & EAGLE-3$^\dagger$ & 5.24 & 4.60 & 3.71 & 3.81 & 4.14 & 4.01 & 2.62 & 2.47 & 2.48 \\
 & DFlash$^\dagger$ & 5.41 & 4.84 & 3.98 & 4.44 & 4.59 & 4.33 & 3.10 & 2.94 & 2.72 \\
 & DSpark ($B{=}7$) & 6.48 & 6.41 & 5.76 & 5.48 & 5.79 & 5.45 & 3.97 & 3.76 & 3.92 \\
 & \textbf{\method} \textbf{($B{=}7$)} & \textbf{7.38} & \textbf{7.30} & \textbf{6.82} & \textbf{6.71} & \textbf{6.93} & \textbf{6.50} & \textbf{5.14} & \textbf{4.92} & \textbf{4.92} \\
 & DSpark ($B{=}16$) & 9.94 & 9.22 & 7.11 & 7.04 & 7.72 & 6.77 & 4.43 & 4.08 & 4.36 \\
 & \textbf{\method} ($B{=}16$) & \textbf{11.61} & \textbf{10.94} & \textbf{9.12} & \textbf{8.81} & \textbf{9.54} & \textbf{8.19} & \textbf{5.72} & \textbf{5.34} & \textbf{5.54} \\
\bottomrule
\end{tabular}
}
\end{table*}

\paragraph{Relative improvement by block size.}
\Cref{tab:gain} reports \emph{absolute} $\tau$ and AR target-only speedup on GSM8K, HumanEval, and MT-Bench.
Across suites, \method consistently raises $\tau$ at $B{=}7$.
At the short block, the GSM8K mean gain over three matched runs is
$+6.1\%$; the HumanEval and MT-Bench gains are
$+14.3\%$ and $+29.3\%$, respectively.
At $B{=}16$, \method raises $\tau$ by $18.6\%$--$28.8\%$ and three-run
mean AR speedup by $7.5\%$--$15.1\%$ relative to matched DSpark on the same
three representative suites.
As \cref{tab:abl_cost} shows, the wider Markov stage costs more per round, but the longer accepted paths reduce the number of proposal rounds enough to improve overall AR speedup.

\begin{table*}[t]
\centering
\caption{Matched gains over DSpark on three representative tasks
(Qwen3-4B, $k{=}4$, $N{=}32$).
All $B{=}16$ entries and GSM8K at $B{=}7$ report three-run means.
Deltas use unrounded measurements.}
\label{tab:gain}
\setlength{\tabcolsep}{2.0pt}
\small
\begin{tabular*}{\textwidth}{@{\extracolsep{\fill}}llccccc@{}}
\toprule
Suite & $B$ & $\tau_{\mathrm{DS}}$ & $\tau_{\mathrm{PC}}$ & $\Delta\tau$ & \shortstack{AR speedup\\(DSpark/\method)} & \shortstack{$\Delta$ AR\\speedup} \\
\midrule
\multirow{2}{*}{GSM8K}
 & $7$ & $6.31$ & $7.24$ & $+0.93$ ($+14.8\%$) & $4.24{\times}/4.50{\times}$ & $+6.1\%$ \\
 & $16$ & $9.41$ & $11.16$ & $+1.75$ ($+18.6\%$) & $6.14{\times}/6.60{\times}$ & $+7.5\%$ \\
\midrule
\multirow{2}{*}{HumanEval}
 & $7$ & $5.60$ & $6.78$ & $+1.18$ ($+21.2\%$) & $3.74{\times}/4.27{\times}$ & $+14.3\%$ \\
 & $16$ & $7.42$ & $9.21$ & $+1.79$ ($+24.1\%$) & $4.89{\times}/5.47{\times}$ & $+11.8\%$ \\
\midrule
\multirow{2}{*}{MT-Bench}
 & $7$ & $3.82$ & $5.07$ & $+1.24$ ($+32.5\%$) & $2.48{\times}/3.20{\times}$ & $+29.3\%$ \\
 & $16$ & $4.28$ & $5.50$ & $+1.23$ ($+28.8\%$) & $2.82{\times}/3.25{\times}$ & $+15.1\%$ \\
\bottomrule
\end{tabular*}
\end{table*}

\paragraph{Scaling across target sizes.}
The acceptance gains in \cref{tab:main} persist from Qwen3-4B to Qwen3-8B
and Qwen3-14B at both block sizes. This consistency indicates that the
benefit is not confined to one target scale. The complete AR-speedup results
for these larger targets are reported in \cref{tab:all_throughput}.

\subsection{Mechanism Isolation: Shared vs.\ Parent-Conditioned Markov Trees}
\label{sec:mechanism}

\Cref{fig:parent-conditioning} illustrates that reusing one position-wise
distribution under different parents can stitch individually likely tokens
into an incoherent path.
We test this failure mode on model outputs using three variants that share
the DSpark checkpoint, prompts, block size $B$, target verifier, and packing
code, thereby separating the effect of branching from that of
parent-specific reconditioning:
\begin{enumerate}[leftmargin=1.4em]
  \item \textbf{DSpark}: the original top-$1$ Markov chain;
  \item \textbf{Shared-Markov tree}: first construct the greedy DSpark
  reference chain using \cref{eq:markov_step}. At each depth, reuse its
  Markov-refined distribution $q_d$ for every tree parent. Thus different
  parents receive the same distribution and child ranking:
  $q_d(\cdot\mid p)=q_d(\cdot\mid p')=q_d(\cdot)$;
  \item \textbf{\method}: following \cref{eq:tree_step}, parent-conditioned
  scoring computes a separate distribution for every concrete frontier
  parent. Parents at the same depth share $L_d$ but may induce different
  child rankings. Parent
  computations within a depth are batched, while the $B$ depth-wise Markov
  stages remain sequential.
\end{enumerate}
These three variants form the controlled mechanism comparison.
The Shared-Markov control retains the jointly trained Markov head but removes
parent-specific reconditioning, isolating the mechanism of interest.
Both tree variants use the same $k$-width frontier, candidate budget,
selection rule, packing implementation, and target verifier.

Separately, we include DFlash+DDTree as an external tree reference under the
same target, GSM8K evaluation set, decoding protocol, and $N{=}32$ budget. Because it
uses a different draft architecture and checkpoint, it provides system-level
context rather than a causal mechanism control. We use the official Qwen3-4B
DFlash $B{=}16$ checkpoint
\href{https://huggingface.co/z-lab/Qwen3-4B-DFlash-b16/tree/main}
{available on HuggingFace}.

\begin{figure*}[t]
  \centering
  \IfFileExists{figures/fig4_mechanism_survival.pdf}{
    \includegraphics[width=0.67\textwidth,trim={0 6pt 0 0},clip]{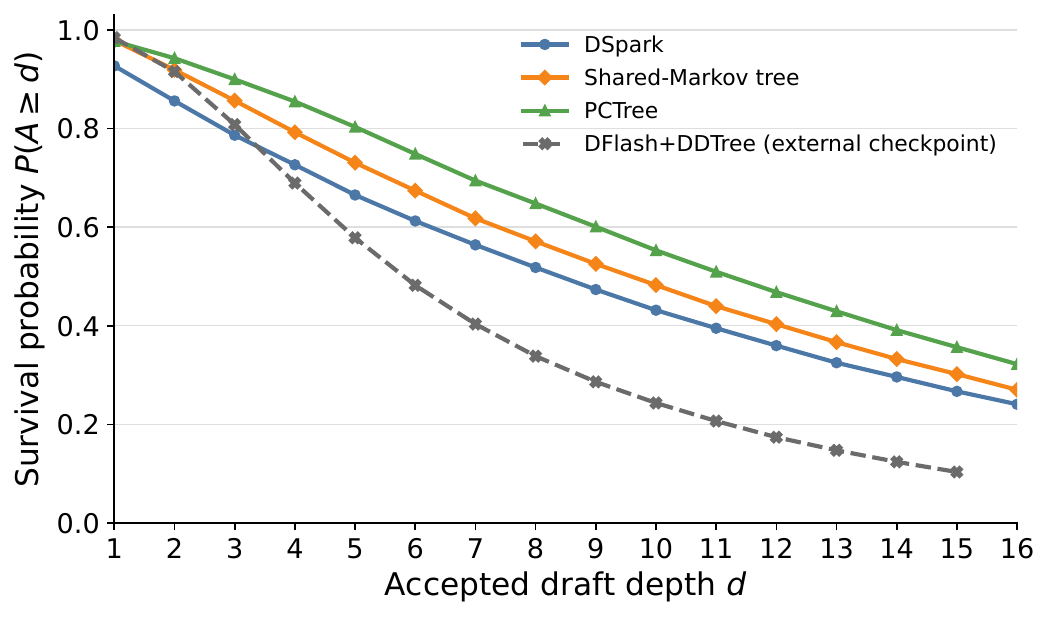}
  }{
    \fbox{\parbox[c][0.20\textheight][c]{0.64\textwidth}{
      \centering
      \Large\textbf{Core Mechanism Result Placeholder}\\[0.8em]
      \normalsize Accepted-depth survival: DSpark, Shared-Markov, \method, and DDTree\\[0.5em]
      \texttt{figures/fig4\_mechanism\_survival.pdf}\\[0.8em]
      Plot $S(d)=\Pr[A\ge d]$ with bootstrap confidence bands.
    }}
  }
  \caption{\textbf{Accepted-depth survival on GSM8K.}
  Each curve gives $S(d)=\Pr[A\ge d]$.
  The DSpark tree controls share the checkpoint, 32-node budget, and search
  policy; Shared-Markov shares one distribution per depth, while \method
  reconditions on each parent.
  DFlash+DDTree (gray dashed) is an external-checkpoint reference.}
  \label{fig:mechanism-survival}
\end{figure*}

We measure rejection cascades through mean acceptance length, verification
rounds per sample, and accepted-depth survival.  The survival curve
$S(d)=\Pr[A\ge d]$ is the probability that a round accepts at least $d$ draft
tokens.  We also report the fraction of verified non-root nodes committed to
the output path (verified-node utilization); the remaining fraction is
uncommitted.  These fractions describe how the node budget is used, not
wall-clock waste, because a tree deliberately verifies alternative siblings
while committing only one path.
\Cref{fig:mechanism-survival,tab:mechanism} report the corresponding
model-based measurements.

\begin{table*}[t]
\centering
\caption{Mechanism isolation on GSM8K ($B{=}16$, $k{=}4$, $N{=}32$).
The DSpark tree controls share the checkpoint, search, packing, and verifier;
only parent-specific reconditioning differs.
DFlash+DDTree uses an external checkpoint and is a system-level reference.
Utilization is the committed fraction of verified non-root nodes.}
\label{tab:mechanism}
\small
\begin{tabular*}{\textwidth}{@{\extracolsep{\fill}}lcccc@{}}
\toprule
Construction & Mean $\tau$ & \shortstack{Verified-node\\utilization (\%)} &
\shortstack{Uncommitted-node\\fraction (\%)} & Rounds/sample \\
\midrule
\multicolumn{5}{l}{\emph{Within-checkpoint DSpark controls}}\\
DSpark & 9.410 & 53.78 & 46.22 & 26.812 \\
Shared-Markov tree & 10.225 & 29.88 & 70.12 & 24.718 \\
\textbf{\method} & \textbf{11.156} & 32.90 & 67.10 & \textbf{22.632} \\
\midrule
\multicolumn{5}{l}{\emph{External tree baseline (different draft checkpoint)}}\\
DFlash+DDTree ($B{=}16$) & 7.485 & 20.92 & 79.08 & 33.826 \\
\bottomrule
\end{tabular*}
\end{table*}

\paragraph{Mechanism result.}
Parent-specific reconditioning improves proposal quality under the same
DSpark checkpoint and 32-node budget.
Relative to Shared-Markov, \method increases mean acceptance length from
$10.225$ to $11.156$ ($+9.1\%$), raises verified-node utilization from
$29.88\%$ to $32.90\%$, and reduces the required rounds per sample from
$24.718$ to $22.632$ ($-8.4\%$).
At depths 8 and 12, the corresponding survival probabilities increase from
$57.1\%$ to $64.8\%$ and from $40.3\%$ to $46.8\%$, respectively.
The survival curves in \cref{fig:mechanism-survival} connect this result to
the distribution-stitching example in \cref{fig:parent-conditioning}.
Because the two controls differ only in reconditioning, the acceptance and
round-count differences measure the effect of conditioning on the concrete
parent.

\paragraph{External tree reference.}
The DFlash+DDTree curve ends at $d{=}15$ because its official $B{=}16$
convention uses one fixed root position and 15 masked future positions.
Its results provide external context, but differences from the DSpark
variants cannot be attributed solely to tree scoring because the draft
architecture and checkpoint also change.

\subsection{Ablation Studies}
\label{sec:ablations}

All ablations use matched large-block drafts at $B{=}16$ and report GSM8K.
The branching-factor sweep uses Qwen3-4B as a representative target.
We focus on $B{=}16$, where the larger candidate space makes branching choices most consequential.

\subsubsection{Branching factor $k$}
\label{sec:abl_k}
\Cref{tab:abl_k} varies $k\in\{1,2,4,8\}$ at fixed $N{=}32$ on $B{=}16$.
The $k{=}1$ column is the chain endpoint (equivalent to DSpark; not a separate baseline);
gains from tree drafting appear as $k$ increases.
On Qwen3-4B, increasing $k$ from $1$ to $2$ yields most of the acceptance and AR-speedup gain.
Performance largely saturates by the default $k{=}4$:
$k{=}8$ provides no further acceptance benefit and changes AR speedup only within timing variability.

\begin{table}[t]
\centering
\caption{Ablation on branching factor $k$ ($B{=}16$, $N{=}32$, GSM8K).
$k{=}1$ $\equiv$ DSpark (chain); $k{\ge}2$ are tree policies.
The shared $k{=}1,4$ endpoints use the three-run main measurements.}
\label{tab:abl_k}
\small
\begin{tabular}{llcccc}
\toprule
Target & Metric & $1$ ($\equiv$ DSpark) & $2$ & $4$ & $8$ \\
\midrule
\multirow{2}{*}{Qwen3-4B}
 & Mean $\tau$ & $9.41$ & $10.85$ & $\mathbf{11.16}$ & $11.14$ \\
 & AR speedup & $6.14{\times}$ & $6.58{\times}$ & $\mathbf{6.60{\times}}$ & $6.59{\times}$ \\
\bottomrule
\end{tabular}
\end{table}

\subsubsection{Verification budget $N$}
\label{sec:abl_n}
\begin{figure*}[t]
  \centering
  \IfFileExists{figures/fig5_budget_scaling.pdf}{
    \includegraphics[width=\textwidth,trim={0 6pt 0 0},clip]{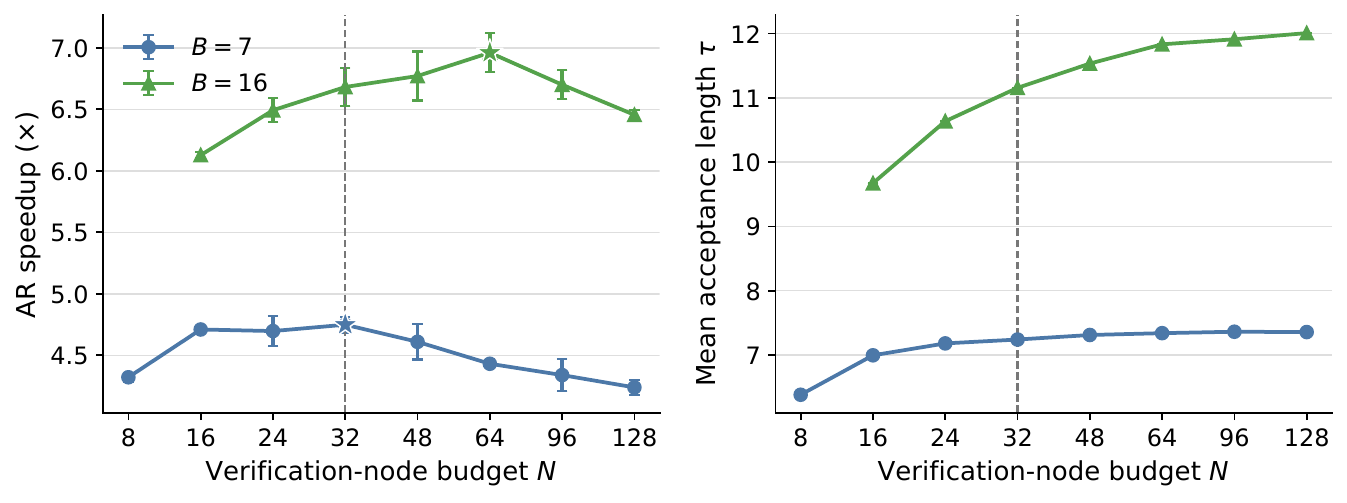}
  }{
    \fbox{\parbox[c][0.25\textheight][c]{0.96\textwidth}{
      \centering
      \Large\textbf{Verification-Budget Scaling Placeholder}\\[0.8em]
      \normalsize Sweep $N\in\{8,16,24,32,48,64,96,128\}$ at $k{=}4$ for $B\in\{7,16\}$\\[0.5em]
      \texttt{figures/fig5\_budget\_scaling.pdf}\\[0.8em]
      Left: AR speedup vs.\ $N$. Right: mean acceptance length vs.\ $N$.
    }}
  }
  \caption{\textbf{Scaling with verification-node budget on GSM8K}
  (Qwen3-4B, $k{=}4$).
  Points are three-run means; AR-speedup error bars show one standard
  deviation.
  The dashed line marks the main setting $N{=}32$, and stars mark the
  highest measured speedup.
  This sweep was timed independently from the main tables.}
  \label{fig:budget-scaling}
\end{figure*}

\Cref{fig:budget-scaling} separates acceptance gains from their end-to-end
cost by sweeping $N$ at fixed $k{=}4$.
Increasing $N$ generally lengthens the accepted prefix, but the improvement
gradually saturates.
AR speedup is therefore non-monotonic: it peaks at $N{=}32$ for $B{=}7$ and
at $N{=}64$ for $B{=}16$, then declines as the extra tree-construction and
verification work outweighs the small remaining acceptance gain.
We retain the pre-specified $N{=}32$ in the main experiments because it is a
strong common operating point across block sizes, rather than selecting a
task-specific budget from GSM8K test performance.
The sweep-specific optima should be interpreted as
hardware- and workload-dependent operating points, not as retroactive tuning
of the main comparisons.

\subsubsection{Cost breakdown}
\label{sec:abl_cost}
\Cref{tab:abl_cost} decomposes average per-round latency on GSM8K.
Because \method changes only candidate expansion, its backbone cost matches
DSpark at each block size. Branching raises the Markov+tree cost, but longer
accepted paths reduce the average number of rounds per sample: from $39.9$ to
$35.1$ at $B{=}7$ and from $26.8$ to $22.6$ at $B{=}16$.
The reduction in proposal rounds offsets the added draft work and improves
AR speedup at both block sizes.
The Markov+tree measurement includes selection, pruning, bookkeeping, and
packing in addition to the Markov projection; consequently, $k{=}4$ does not
imply exactly $4\times$ DSpark's batch-$1$ Markov latency.

\begin{table*}[t]
\centering
\caption{Phase cost breakdown on GSM8K
(Qwen3-4B, $k{=}4$, $N{=}32$).
Phase latency is wall time per proposal round.
AR speedups are three-run means; phase latencies are representative matched
measurements.}
\label{tab:abl_cost}
\small
\begin{tabular*}{\textwidth}{@{\extracolsep{\fill}}llccccc@{}}
\toprule
& & & & \multicolumn{2}{c}{Avg.\ phase latency (ms/rd)} & \\
\cmidrule(lr){5-6}
$B$ & Method & Mean $\tau$ & Avg.\ rounds/sample &
Backbone & Markov+tree & AR speedup \\
\midrule
\multirow{2}{*}{$7$}
 & DSpark & $6.31$ & $39.9$ & $3.46$ & $0.65$ & $4.24{\times}$ \\
 & \textbf{\method} & $7.24$ & $35.1$ & $3.46$ & $3.52$ & $\mathbf{4.50{\times}}$ \\
\midrule
\multirow{2}{*}{$16$}
 & DSpark & $9.41$ & $26.8$ & $3.55$ & $1.09$ & $6.14{\times}$ \\
 & \textbf{\method} & $11.16$ & $22.6$ & $3.55$ & $5.24$ & $6.60{\times}$ \\
\bottomrule
\end{tabular*}
\end{table*}

\section{Limitations}
\label{sec:limitations}

There are some limitations in this work. (1) Tree quality remains bounded by the pretrained draft distributions; a training-free expansion policy cannot fix a systematically weak Markov head.
(2) Layer-wise top-$k$ pruning is a heuristic and may drop a globally optimal path under the node budget $N$.
(3) Very large $k$ or $N$ can make draft construction and verify attention non-negligible on small targets or memory-tight devices.
(4) Although AR speedup improves in the present BF16 runs, the gain depends on hardware, precision, attention kernels, and the tree budget.

\section{Conclusion}
\label{sec:conclusion}

We presented \method, a training-free method that turns DSpark's semi-autoregressive chain drafts into budgeted draft trees.
By expanding top-$k$ Markov children on shared backbone logits and verifying the resulting tree in one target forward, \method increases acceptance length without any retraining.
AR-speedup measurements suggest that the acceptance gains can translate into end-to-end improvements at $B{=}7$ and $B{=}16$ across Qwen3-\{4B,8B,14B\}.
More generally, a drafter that factorizes into a parallel block and a cheap sequential refinement can support tree verification through an inference-policy change. The benefit is largest when early rejection makes long linear blocks inefficient.

\bibliographystyle{plainnat}
\bibliography{references}

\clearpage
\appendix
\section{Full Experimental Protocol}
\label{app:setup}
\label{app:protocol}

\paragraph{Prompts and decoding.}
We apply the target model's chat template with thinking disabled and evaluate
the first user turn.
Decoding is greedy with batch size one, a 512-token output cap, and the
model's EOS stopping rule.
DSpark and \method use identical prompts, ordering, and stopping
criteria.
The pre-specified reference tree uses $k{=}4$ and $N{=}32$ including the root; confidence scheduling and near-tie re-verification are disabled.
The target and draft models, their KV caches, and hidden states use bfloat16.

\paragraph{Evaluation data.}
Table~\ref{tab:data-manifest} records the upstream releases used by our
evaluation harness.
\begin{table*}[t]
\centering
\caption{Evaluation datasets and their upstream releases. ``Rows'' denotes
the size of the corresponding evaluation file.}
\label{tab:data-manifest}
\small
\begin{tabular*}{\textwidth}{@{\extracolsep{\fill}}llr}
\toprule
Dataset & Upstream release & Rows \\
\midrule
GSM8K & \href{https://huggingface.co/datasets/openai/gsm8k}{\texttt{openai/gsm8k}} & 1,319 \\
MATH-500 & \href{https://huggingface.co/datasets/HuggingFaceH4/MATH-500}{\texttt{HuggingFaceH4/MATH-500}} & 500 \\
AIME25 & \href{https://huggingface.co/datasets/MathArena/aime_2025}{\texttt{MathArena/aime\_2025}} & 30 \\
MBPP & \href{https://huggingface.co/datasets/google-research-datasets/mbpp}{\texttt{google-research-datasets/mbpp}} & 257 \\
HumanEval & \href{https://huggingface.co/datasets/openai/openai_humaneval}{\texttt{openai/openai\_humaneval}} & 164 \\
LiveCodeBench & \href{https://huggingface.co/datasets/livecodebench/code_generation_lite}{\texttt{livecodebench/code\_generation\_lite}} & 1,055 \\
MT-Bench & \href{https://huggingface.co/datasets/HuggingFaceH4/mt_bench_prompts}{\texttt{HuggingFaceH4/mt\_bench\_prompts}} & 80 \\
Alpaca & \href{https://huggingface.co/datasets/tatsu-lab/alpaca}{\texttt{tatsu-lab/alpaca}} & 52,002 \\
Arena-Hard & \href{https://huggingface.co/datasets/lmarena-ai/arena-hard-auto}{\texttt{lmarena-ai/arena-hard-auto}} & 750 \\
\bottomrule
\end{tabular*}
\end{table*}

\paragraph{Software and hardware.}
Experiments run on one NVIDIA H20 in bfloat16 using PyTorch 2.11,
Transformers 5.5, CUDA 12.8, and SDPA attention.
End-to-end timing excludes model and dataset loading but includes prompt
processing, data transfer, and generation.

\paragraph{Warmup and repeated timing.}
We perform three matched full-run repetitions for all Qwen3-4B $B{=}16$
main configurations and for Qwen3-4B GSM8K at $B{=}7$.
Other main-table configurations currently have one run; each configuration
in the verification-budget sweep is also repeated three times.
Model and dataset loading occur before timing, but no prompt-level warmup is
used, so the first prompt is included in the measured loop.
The dedicated $B{=}7$ repeatability run alternates DSpark/\method execution
order; GPU persistence mode and application clocks remain at the host defaults.
Each run reports total generated tokens divided by end-to-end wall time, and
AR speedup is computed within each repetition before reporting
mean$\pm$standard deviation.

\paragraph{Checkpoints.}
Target and draft checkpoint identifiers:
\begin{itemize}[leftmargin=1.2em]
  \item Target models:
  \href{https://huggingface.co/Qwen/Qwen3-4B}{\texttt{Qwen/Qwen3-4B}},
  \href{https://huggingface.co/Qwen/Qwen3-8B}{\texttt{Qwen/Qwen3-8B}}, and
  \href{https://huggingface.co/Qwen/Qwen3-14B}{\texttt{Qwen/Qwen3-14B}};
  \item Official short-block DSpark drafts ($B{=}7$) for Qwen3-4B,
  Qwen3-8B, and Qwen3-14B:
  \href{https://huggingface.co/deepseek-ai/dspark_qwen3_4b_block7}
  {\texttt{4B}},
  \href{https://huggingface.co/deepseek-ai/dspark_qwen3_8b_block7}
  {\texttt{8B}}, and
  \href{https://huggingface.co/deepseek-ai/dspark_qwen3_14b_block7}
  {\texttt{14B}};
  \item Qwen3-\{4B,8B,14B\} + internally trained large-block DSpark drafts
  ($B{=}16$): separate target-matched checkpoints trained with the official
  DSpark recipe on
  \href{https://huggingface.co/datasets/mlabonne/open-perfectblend}
  {\texttt{mlabonne/open-perfectblend}} after regenerating assistant
  responses with the corresponding target model.
  \item DFlash+DDTree external baseline: official Qwen3-4B DFlash
  $B{=}16$ checkpoint
  \href{https://huggingface.co/z-lab/Qwen3-4B-DFlash-b16/tree/main}
  {\texttt{z-lab/Qwen3-4B-DFlash-b16}}.
\end{itemize}

\paragraph{Tensor implementation.}
Selected tree nodes are stored in a flat token tensor.
Each node's position identifier is its tree depth plus the verified prefix length, so siblings share a position while retaining distinct ancestry.
The tree mask permits a node to attend to the verified prefix, itself, and
its root-to-node ancestors.
Retrieve indices enumerate padded root-to-leaf rows used to compare each proposed child against the preceding target logit; after selecting the longest accepted row, KV tensors and target hidden states are gathered along that row and compacted after the verified prefix.

\section{Complete AR-Speedup Results}
\label{app:throughput}

\begin{table*}[h]
\centering
\caption{End-to-end AR speedup for all nine tasks.
Each pair uses a matched DSpark/\method harness.
Qwen3-4B $B{=}16$ and Qwen3-4B GSM8K at $B{=}7$ report three-run means.}
\label{tab:all_throughput}
\small
\resizebox{\textwidth}{!}{%
\begin{tabular}{lcccccccccccc}
\toprule
& \multicolumn{2}{c}{4B, $B{=}7$} & \multicolumn{2}{c}{4B, $B{=}16$} &
\multicolumn{2}{c}{8B, $B{=}7$} & \multicolumn{2}{c}{8B, $B{=}16$} &
\multicolumn{2}{c}{14B, $B{=}7$} & \multicolumn{2}{c}{14B, $B{=}16$} \\
\cmidrule(lr){2-3}\cmidrule(lr){4-5}\cmidrule(lr){6-7}\cmidrule(lr){8-9}\cmidrule(lr){10-11}\cmidrule(lr){12-13}
Suite & DSpark & \textbf{\method} & DSpark & \textbf{\method} & DSpark & \textbf{\method} & DSpark & \textbf{\method} & DSpark & \textbf{\method} & DSpark & \textbf{\method} \\
\midrule
GSM8K & $4.24{\times}$ & $\mathbf{4.50{\times}}$ & $6.14{\times}$ & $\mathbf{6.60{\times}}$ & $4.43{\times}$ & $\mathbf{4.86{\times}}$ & $6.66{\times}$ & $\mathbf{7.09{\times}}$ & $4.28{\times}$ & $\mathbf{4.88{\times}}$ & $6.93{\times}$ & $\mathbf{7.21{\times}}$ \\
MATH-500 & $4.21{\times}$ & $\mathbf{4.34{\times}}$ & $6.02{\times}$ & $\mathbf{6.37{\times}}$ & $4.36{\times}$ & $\mathbf{4.71{\times}}$ & $6.26{\times}$ & $\mathbf{6.63{\times}}$ & $4.37{\times}$ & $\mathbf{4.91{\times}}$ & $6.20{\times}$ & $\mathbf{6.76{\times}}$ \\
AIME25 & $3.65{\times}$ & $\mathbf{4.17{\times}}$ & $4.51{\times}$ & $\mathbf{5.07{\times}}$ & $3.86{\times}$ & $\mathbf{4.34{\times}}$ & $4.71{\times}$ & $\mathbf{5.39{\times}}$ & $3.97{\times}$ & $\mathbf{4.46{\times}}$ & $4.78{\times}$ & $\mathbf{5.46{\times}}$ \\
MBPP & $3.61{\times}$ & $\mathbf{4.06{\times}}$ & $4.67{\times}$ & $\mathbf{5.04{\times}}$ & $3.75{\times}$ & $\mathbf{4.50{\times}}$ & $4.64{\times}$ & $\mathbf{5.58{\times}}$ & $3.75{\times}$ & $\mathbf{4.38{\times}}$ & $4.70{\times}$ & $\mathbf{5.55{\times}}$ \\
HumanEval & $3.74{\times}$ & $\mathbf{4.27{\times}}$ & $4.89{\times}$ & $\mathbf{5.47{\times}}$ & $4.17{\times}$ & $\mathbf{4.60{\times}}$ & $5.45{\times}$ & $\mathbf{5.83{\times}}$ & $4.08{\times}$ & $\mathbf{4.69{\times}}$ & $5.40{\times}$ & $\mathbf{5.62{\times}}$ \\
LCB & $3.49{\times}$ & $\mathbf{3.74{\times}}$ & $4.34{\times}$ & $\mathbf{4.51{\times}}$ & $3.61{\times}$ & $\mathbf{4.01{\times}}$ & $4.30{\times}$ & $\mathbf{4.58{\times}}$ & $3.45{\times}$ & $\mathbf{3.75{\times}}$ & $4.07{\times}$ & $\mathbf{4.36{\times}}$ \\
MT-Bench & $2.48{\times}$ & $\mathbf{3.20{\times}}$ & $2.82{\times}$ & $\mathbf{3.25{\times}}$ & $2.60{\times}$ & $\mathbf{3.34{\times}}$ & $3.00{\times}$ & $\mathbf{3.50{\times}}$ & $2.74{\times}$ & $\mathbf{3.42{\times}}$ & $3.10{\times}$ & $\mathbf{3.47{\times}}$ \\
Alpaca & $2.39{\times}$ & $\mathbf{3.02{\times}}$ & $2.69{\times}$ & $\mathbf{3.09{\times}}$ & $2.65{\times}$ & $\mathbf{3.26{\times}}$ & $2.76{\times}$ & $\mathbf{3.37{\times}}$ & $2.61{\times}$ & $\mathbf{3.38{\times}}$ & $2.89{\times}$ & $\mathbf{3.40{\times}}$ \\
Arena-Hard & $2.52{\times}$ & $\mathbf{2.93{\times}}$ & $2.86{\times}$ & $\mathbf{3.16{\times}}$ & $2.67{\times}$ & $\mathbf{3.12{\times}}$ & $2.98{\times}$ & $\mathbf{3.38{\times}}$ & $2.64{\times}$ & $\mathbf{3.16{\times}}$ & $2.91{\times}$ & $\mathbf{3.30{\times}}$ \\
\bottomrule
\end{tabular}%
}
\end{table*}

\section{Repeatability Check}
\label{app:repeatability}

We repeat the Qwen3-4B $B{=}16$ main evaluation across all nine tasks.
We also repeat GSM8K at $B{=}7$ because it had the smallest AR-speedup margin
in the initial evaluation and was therefore the most susceptible to timing
noise.
Within each matched repetition, DSpark and \method use the same
checkpoint, evaluation inputs in the same order, generated-token cap, decoding
settings, and execution environment.
\Cref{tab:repeatability} reports the GSM8K runs individually; the main tables
retain the more compact three-run means.

\begin{table}[h]
\centering
\caption{Repeatability of Qwen3-4B GSM8K
($k{=}4$, $N{=}32$).
Mean$\pm$standard deviation is computed over three full runs.}
\label{tab:repeatability}
\setlength{\tabcolsep}{3.5pt}
\small
\begin{tabular}{ccccc}
\toprule
$B$ & Run & DSpark & \method & Gain \\
\midrule
\multirow{4}{*}{$7$}
 & 1 & $4.31{\times}$ & $4.54{\times}$ & $+5.4\%$ \\
 & 2 & $4.22{\times}$ & $4.47{\times}$ & $+5.8\%$ \\
 & 3 & $4.18{\times}$ & $4.48{\times}$ & $+7.1\%$ \\
 & Mean$\pm$std. & $4.24{\pm}0.07{\times}$ & $\mathbf{4.50{\pm}0.04{\times}}$ & $\mathbf{+6.1{\pm}0.9\%}$ \\
\midrule
\multirow{4}{*}{$16$}
 & 1 & $6.09{\times}$ & $6.64{\times}$ & $+9.0\%$ \\
 & 2 & $6.22{\times}$ & $6.57{\times}$ & $+5.7\%$ \\
 & 3 & $6.10{\times}$ & $6.57{\times}$ & $+7.8\%$ \\
 & Mean$\pm$std. & $6.14{\pm}0.07{\times}$ & $\mathbf{6.60{\pm}0.04{\times}}$ & $\mathbf{+7.5{\pm}1.7\%}$ \\
\bottomrule
\end{tabular}
\end{table}

\section{Reproducibility Statement}
\label{app:reproducibility}
The draft factorization, search and selection rules, verification packing,
decoding parameters, software stack, checkpoints, and data sources are
documented above.
The official public checkpoints used in our evaluation are linked above.
The $B{=}16$ drafts are internally trained using the procedure described in
\cref{app:setup}.

\end{document}